\pdfoutput=1
\documentclass[11pt]{article}
\usepackage[table,dvipsnames]{xcolor}
\usepackage{acl}
\usepackage{times}
\usepackage{latexsym}
\usepackage[T1]{fontenc}
\usepackage[utf8]{inputenc}
\usepackage{microtype}
\usepackage{inconsolata}
\usepackage{amsmath}
\usepackage{amssymb}
\usepackage{graphicx}
\usepackage{enumerate}
\usepackage{balance}
\usepackage{blindtext}
\usepackage{booktabs}
\usepackage{colortbl}
\usepackage{float}
\usepackage{makecell}
\usepackage{multicol}
\usepackage{multirow}
\usepackage{tabularx}
\usepackage{marvosym}
\usepackage{latexsym}
\usepackage[T1]{fontenc}
\usepackage[utf8]{inputenc}
\usepackage{microtype}
\usepackage{inconsolata}
\usepackage{amsmath}
\usepackage{amssymb}
\usepackage{graphicx}
\usepackage{enumerate}
\usepackage{balance}
\usepackage{blindtext}
\usepackage{listings}
\usepackage{booktabs}
\usepackage{colortbl}
\usepackage{tcolorbox}
\usepackage{marvosym}
\usepackage{float}
\usepackage{makecell}
\usepackage{multicol}
\usepackage{multirow}
\usepackage{tabularx}
\usepackage{marvosym}
\usepackage{stfloats}
\usepackage{graphicx}
\usepackage{subfigure}
\usepackage{subcaption}
\usepackage{ulem}
\lstdefinestyle{PythonCode}{
    language=Python,
    basicstyle=\ttfamily,
    breaklines=true,
    keywordstyle=\bfseries\color{NavyBlue},
    morekeywords={},
    emph={self},
    emphstyle=\bfseries\color{Rhodamine},
    commentstyle=\itshape\color{black!50!white},
    stringstyle=\bfseries\color{PineGreen!90!black},
    columns=flexible,
}

\lstdefinestyle{BashCode}{
    language=Bash,
    basicstyle=\ttfamily\color{white}, 
    backgroundcolor=\color{black},     
    breaklines=true,
    keywordstyle=\bfseries\color{MidnightBlue},
    morekeywords={},
    emph={},
    emphstyle=\bfseries\color{Purple},
    commentstyle=\itshape\color{black!50!white},
    stringstyle=\bfseries\color{OliveGreen!90!black},
    columns=flexible,
}

\newcommand{\ie}{\textit{i.e., }}
\newcommand{\eg}{\textit{e.g., }}

\newcommand\footnoteONLYtext[1]{
    \let \mybackup \thefootnote
    \let \thefootnote \relax
    \footnotetext{#1}
    \let \thefootnote \mybackup
    \let \mybackup \imareallyundefinedcommand}

\title{Experiential Co-Learning of Software-Developing Agents}

\author{
  \textbf{Chen Qian}{\footnotesize $^{\dagger\bigstar}$} \quad
  \textbf{Yufan Dang}{\footnotesize $^{\dagger\bigstar}$} \quad 
  \textbf{Jiahao Li}{\footnotesize $^\spadesuit$} \quad 
  \textbf{Wei Liu}{\footnotesize $^\bigstar$} \quad 
  \textbf{Zihao Xie}{\footnotesize $^\bigstar$} \quad 
  \\
  \textbf{Yifei Wang}{\footnotesize $^\bigstar$} \quad
  \textbf{Weize Chen}{\footnotesize $^\bigstar$} \quad 
  \textbf{Cheng Yang}{\footnotesize $^\clubsuit$\textsuperscript{\Letter}} \quad 
  \textbf{Xin Cong}{\footnotesize $^\bigstar$} \quad 
  \textbf{Xiaoyin Che}{\footnotesize $^\blacklozenge$} \quad 
  \\
  \textbf{Zhiyuan Liu}{\footnotesize $^{\bigstar}$\textsuperscript{\Letter}} \quad  
  \textbf{Maosong Sun}{\footnotesize $^{\bigstar}$\textsuperscript{\Letter}} \\
  {\footnotesize $^\bigstar$}Tsinghua University \quad
  {\footnotesize $^{\spadesuit}$}Dalian University of Technology \\
  {\footnotesize $^\clubsuit$}Beijing University of Posts and Telecommunications \quad
  {\footnotesize $^\blacklozenge$}Siemens \\
  \href{qianc62@gmail.com}{\texttt{qianc62@gmail.com}} \quad 
  \href{dangyf21@mails.tsinghua.edu.cn}{\texttt{dangyf21@mails.tsinghua.edu.cn}} \quad \\
  \href{yangcheng@bupt.edu.cn}{\texttt{yangcheng@bupt.edu.cn}} \quad 
  \href{liuzy@tsinghua.edu.cn}{\texttt{liuzy@tsinghua.edu.cn}} \quad 
  \href{sms@tsinghua.edu.cn}{\texttt{sms@tsinghua.edu.cn}}
}

\begin{document}

\maketitle

\footnoteONLYtext{$^\dagger$Equal Contribution.}
\footnoteONLYtext{$^{\text{\Letter}}$Corresponding Author.}

\begin{abstract}
Recent advancements in large language models (LLMs) have brought significant changes to various domains, especially through LLM-driven autonomous agents.
A representative scenario is in software development, where LLM agents demonstrate efficient collaboration, task division, and assurance of software quality, markedly reducing the need for manual involvement.
However, these agents frequently perform a variety of tasks independently, without benefiting from past experiences, which leads to repeated mistakes and inefficient attempts in multi-step task execution.
To this end, we introduce \textit{Experiential Co-Learning}, a novel LLM-agent learning framework in which instructor and assistant agents gather shortcut-oriented experiences from their historical trajectories and use these past experiences for future task execution.
The extensive experiments demonstrate that the framework enables agents to tackle unseen software-developing tasks more effectively.
We anticipate that our insights will guide LLM agents towards enhanced autonomy and contribute to their evolutionary growth in cooperative learning.
The code and data are available at \url{https://github.com/OpenBMB/ChatDev}.
\end{abstract}

\section{Introduction}
In the ever-evolving field of artificial intelligence, large language models (LLMs) have marked a transformative shift across numerous domains \cite{vaswani2017attention,NEURIPS2020_1457c0d6,bubeck2023sparks}.
Despite their impressive abilities, when dealing with complex situations that extend beyond mere chatting, these models show certain limitations inherent in their standalone capabilities~\cite{AutoGPT}.
Recent research in \textit{autonomous agents} has significantly advanced LLMs by integrating sophisticated features like context-sensitive memory~\cite{park2023generative}, multi-step planning~\cite{wei2022chain}, and strategic tool use~\cite{schick2023toolformer}.
This enhancement has expanded their capacity to effectively manage a broader spectrum of complex tasks, including social simulation~\cite{park2023generative,wang2023humanoid,hua2023war}, software development~\cite{GPTEngineer,qian2023communicative}, game playing~\cite{wang2023voyager,zhu2023ghost,wang2023avalons,gong2023mindagent}, and scientific research~\cite{huang2023benchmarking,LLM-Research-Feedback-2023}.
In order to study the cooperative dynamics of autonomous agents more pertinently, we choose software development~\cite{mills1976software} as an representative scenario, due to its \textit{complexity} that demands a blend of natural and programming language skills~\cite{mills1976software}, the \textit{processuality} that often requires an in-depth understanding of coding and continuous alterations~\cite{barki1993toward}, and the \textit{objectivity} of code that can provide quantifiable feedback~\cite{compton2002reconfigurable}.

With the development of autonomous-agent technology, a  successful breakthrough has been the integration of communication among multiple agents \cite{park2023generative,li2023camel,qian2023communicative}.
Representative methods epitomize this methodology by segmenting task execution into distinct subtasks.
Through engaging in cooperative communication, agents participate in instructive or responsive conversations, collaboratively contributing to the achievement of a cohesive and automated solution for task execution.
For example, in ChatDev~\cite{qian2023communicative}, a recent agent-communication framework for software development, a reviewer agent in charge of an external compiler iteratively provides instructions for software optimization (\eg completing unimplemented code, making functional changes, and debugging programs), to which a programmer agent then reacts to these instructions by appropriately updating the source code.
The development of a more adaptive and proactive approach to task-solving by these agents marks a significant leap in autonomy, going beyond the typical prompt-guided dynamic in human-computer communications~\cite{yang2023large} and substantially reducing dependence on human involvement~\cite{li2023camel,qian2023communicative,wu2023autogen}.

However, when confronted with a diverse range of task types, current multi-agent collaboration methods tend to handle each task independently, which largely stems from the absence of a methodology that can effectively incorporate the experiences accumulated from previously completed tasks~\cite{qian2023communicative,chen2023agentverse,park2023generative,hong2023metagpt}.
Consequently, the inexperience nature frequently results in repetitive errors or unnecessary trial-and-error processes through multi-step task execution, ultimately necessitating additional human involvement especially when these methods are applied to real-world scenarios.

How do we design, gather, and apply useful experiences to enhance multi-agent collaboration?
In this paper, we propose \textit{Experiential Co-Learning}, a novel multi-agent learning paradigm designed to boost agents' software-developing abilities through the utilization of experiences gathered from their historical communications.
The method regards agents into two functional roles—instructor and assistant, involving three core modules:
1) the \textit{co-tracking} module promotes communicative rehearsals between the agents, fostering cooperative exploration and the creation of procedural trajectories for various training tasks;
2) the \textit{co-memorizing} module heuristically mines "shortcuts"\footnote{The term "shortcut" is used positively to denote a more efficient pathway, unlike in some papers where it implies a superficial correlation~\cite{geirhos2020shortcut}.} from historical trajectories using external environment feedback, which are then preserved in their experience pools in an interleaved manner;
3) the \textit{co-reasoning} module encourages agents to enhance their instructions and solutions by utilizing their collective experience pools when facing unseen tasks.
The comprehensive assessment of collaborative processes between autonomous agents in diverse software development tasks reveals that the proposed framework significantly boosts collaborative efficiency and reduces the need for extra human involvement.

In summary, our main contributions include:
\begin{enumerate}[$\bullet$]
\item To our knowledge, this study is the first to integrate past experiences into the LLM-powered multi-agent collaboration. Through co-tracking, co-memorizing, and co-reasoning, this framework facilitates cooperative learning among two distinct agent types (instructor and assistant) by leveraging heuristical experiences extracted from their historical task execution trajectories.
\item We propose to construct task execution graphs based on procedural trajectories, in which "shortcuts" linking non-adjacent nodes are extracted as experiences, which can effectively motivate agents to engage in shortcut thinking during reasoning.
\item We conducted extensive experiments from multiple perspectives to validate the effectiveness of our framework. The findings highlight the enhanced quality and efficiency of agents' collaborative behavior in software development.
\end{enumerate}

\section{Related Work}
Trained on vast datasets to comprehend and manipulate billions of parameters, LLMs have become pivotal in natural language processing~\cite{NEURIPS2020_1457c0d6,bubeck2023sparks,vaswani2017attention,radford2019language,touvron2023llama,wei2022emergent,Shanahan2023,chen2021evaluating,brants2007large,chen2021evaluating,ouyang2022training,yang2023large,qin2023large,kaplan2020scaling}. Recent progress, particularly in the filed of autonomous agents~\cite{zhou2023webarena,wang2023voyager,park2023generative,wang2023humanoid,AutoGPT,GPTEngineer,wang2023promptagent}, is largely attributed to the foundational advances in LLMs. These agents utilize the robust capabilities of LLMs, displaying remarkable skills in memory~\cite{park2023generative,sumers2024cognitive}, planning~\cite{GPTEngineer,chen2023agentverse,liu2023bolaa} and tool use~\cite{schick2023toolformer,cai2023large,qin2023toolllm,ruan2023tptu,GPT4Tools}, enabling them to operate independently in complex, real-world scenarios~\cite{zhao2023expel,zhou2023webarena,ma2023laser,zhang2023generative,wang2023large,ding2023designgpt,weng2023prompt}.
Since agents like Reflexion~\cite{shinn2023reflexion} showcase feedback-based improvement yet often lack cross-task experience, to enable an autonomous agent to improve through trial and error across various training tasks, ExpeL~\cite{zhao2023expel} innovatively defines experience as a history of successful task trajectories and utilizes these experiences for in-context reasoning.

Parallel to these attempts, the autonomous communication among multiple agents is now a promising paradigm, signaling a shift towards multi-agent collaboration paradigm~\cite{park2023generative,zhou2023agents,chen2023agentverse,chan2023chateval,chen2023gamegpt,Cohen2023LMVL,li2023metaagents,hua2023war, guo2024large}.
Among them, software-developing agents facilitate the breakdown of complex tasks into more manageable, finer-grained subtasks~\cite{hong2023metagpt,qian2023communicative}.
An instructor issues directional instructions, and an assistant provides relevant solutions, facilitating a streamlined workflow for task execution. 
This approach not only boosts productivity but also exhibits a degree of quality that surpasses the traditional prompt-guided paradigm in human-computer communications, reducing the need for manual involvement~\cite{li2023camel,chen2023agentverse}.

\begin{figure*}[thb]
    \centering
    \includegraphics[width=1\textwidth]{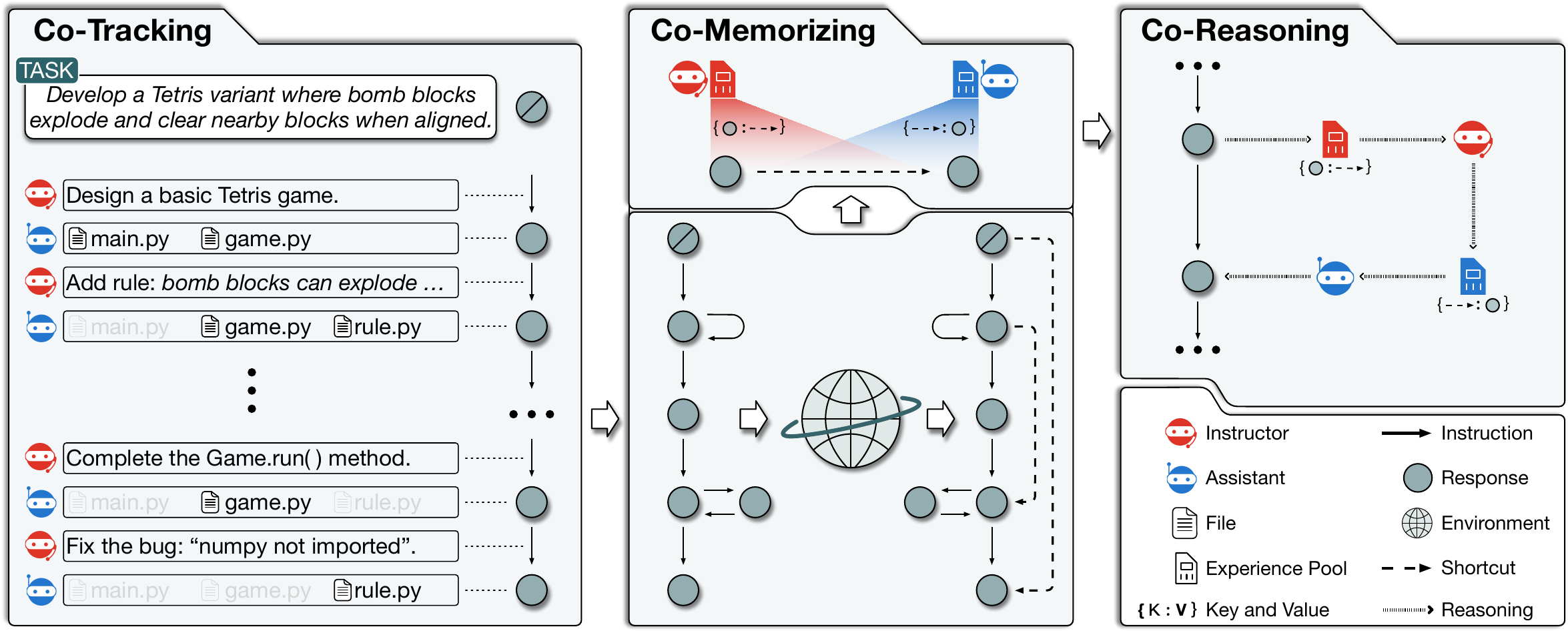}
    \caption{The framework of Experiential Co-Learning. The co-tracking module promotes communicative rehearsals between the agents, fostering cooperative exploration and the creation of procedural trajectories for various training tasks. The co-memorizing module heuristically extracts "shortcuts" from the trajectories under external supervision, integrating these heuristic shortcuts into their collective experience pools. The co-reasoning module combines agents' collective experience pools to foster an communication of augmented instructions and solutions, improving their ability to collaboratively solve unseen tasks.}
    \label{fig:framework}
\end{figure*}

\section{Experiential Co-Learning}

Traditional multi-agent collaboration methods often neglect to accumulate experience from past tasks, leading to repetitive errors or unnecessary trial-and-error processes in similar future tasks~\cite{qian2023communicative}. To remedy this, we propose \textit{Experiential Co-Learning}, illustrated in Figure \ref{fig:framework}, powered by two distinct autonomous agents and comprising three essential modules:
1) the \textit{co-tracking} module establishes a rehearsal collaboration between an instructor and an assistant, focusing on tracking their cooperative "procedural trajectories" for various training tasks, showcasing clear strategies in their communicative collaboration;
2) the \textit{co-memorizing} module heuristically extracts "shortcuts" from the trajectories under external supervision, integrating these heuristic shortcuts into their collective experience pools;
3) the \textit{co-reasoning} module combines the two agents' collective experience pools to foster an communication of augmented instructions and solutions, improving their ability to collaboratively solve unseen tasks.

\subsection{Co-Tracking}
The co-tracking module sets up a collaborative task execution between an instructor and an assistant, with the goal of tracking procedural trajectories for various training tasks \cite{qin2023toolllm}. Each trajectory captures the dynamic progression of a specific task, detailing the evolving roadmap and clearly illustrating the interacted solutions throughout the task execution process.

Formally, in the set of training tasks  $\mathcal{T}$, each task $t \in \mathcal{T}$ fosters a functional interplay of communications between agents, streamlined towards the effective execution of the task.
During this procedure, the instructor gives a series of instructions (\(\mathcal{I} = \{i_1, i_2, \cdots, i_n\}\)), to which the assistant responds with a matching sequence of solutions (\(\mathcal{S} = \{s_1, s_2, \cdots, s_n\}\)), where each solution represents an intact software code.
This communicative dynamic can be naturally modeled as a directed chain \( \mathcal{G} = (\mathcal{N}, \mathcal{E}) \):
\begin{equation*}
\begin{aligned}
\mathcal{N}\!&=\!\{ s_j | s_j \in \mathcal{S} \}\!\cup\!\{s_0\} \\
\mathcal{E}\!&=\!\{ (s_j, i_{j+1}, s_{j+1}) | s_j, s_{j+1}\!\in\!\mathcal{S}, i_{j+1}\!\in\!\mathcal{I} \}
\end{aligned}
\end{equation*}
where \( \mathcal{N} \) represents the nodes corresponding to the solutions (with \( s_0 \) denoting the initial, typically empty solution), and \( \mathcal{E} \) denotes the edges corresponding to the instructions. Each edge \( (s_j, i_{j+1}, s_{j+1}) \) illustrates the transition from one solution \( s_j \) to the modified one \( s_{j+1} \), guided by the instruction \( i_{j+1} \).

\subsection{Co-Memorizing}

We observed that not all progressions in the chain (\ie a single round of software optimization) lead to better solutions. This includes \textit{solution backtracking}, where optimization loops back to earlier content, and \textit{correct-to-failure degeneration}, where functioning software is inadvertently changed into a unexecutive solution. These scenarios suggest some steps in the process are redundant or ineffective, indicating that solely memorizing complete historical trajectories may be insufficient for designing agents' experiences.
Thus, we convert the chain into a \textit{task execution graph} to map nodes with the same content in the chain to a shared node in a redefined ($\leftarrow$) graph:
\begin{equation*}
\begin{aligned}
\mathcal{N}\!&\!\leftarrow\!\{ \phi(s_j) | s_j\!\in\!\mathcal{N} \} \\
\mathcal{E}\!&\!\leftarrow\!\{ (\phi(s_j), i_{j+1}, \phi(s_{j+1})) | (s_j, i_{j+1}, s_{j+1})\!\in\!\mathcal{E} \}
\end{aligned}
\end{equation*}
where \(\phi\) is a mapping rule using a hash function~\cite{sasaki2009finding}.
This approach efficiently groups identical solutions and highlights repetitions, serving as a "state transition graph" throughout the task execution process, as visualized in Figure~\ref{fig:framework}.

Furthermore, in a task execution process, each edge linking two adjacent nodes signifies one round of autonomous solution optimization by the agents, indicating that the agents have already possessed the corresponding decision-making capability for each existing edge.
Hence, relying solely on existing edges might not suffice for the design of agents' past experiences.
For this purpose, the co-memorizing module is devised to heuristically identify shortcuts linking non-adjacent nodes on the graph. These form the basis of the agents' experience pools in practical reasoning, with the goal of accelerating future task execution.

\noindent\paragraph{Node Estimation} Firstly, the score of each node \(s_j\) in \(\mathcal{N}\) is estimated using an external feedback signal, calculated in a pairwise manner:
\begin{equation*}
\omega(s_j)\!=\!sim(s_j, task)\!\times\!sim(s_j, s_{|\mathcal{N}|})\!\times\![\![s_j]\!]
\end{equation*}
where \( sim(\cdot,\cdot) \) calculates the similarity between a node with another node or a task requirement, achieved through the use of an external code embedder and a text embedder, while \( [\![\cdot]\!] \) indicates a binary signal indicating whether compilation is successful via an external compiler.
The heuristic scoring rule assigns high evaluations to solutions that meet the task requirements, resemble the final solution, and are validated by an external tool.

\noindent\paragraph{Shortcut Extraction} To discover informative shortcuts, we selectively identify shortcuts linking non-adjacent nodes that exceed an information gain threshold \(\epsilon\):
\begin{equation*}
\begin{aligned}
\mathsf{S}\!=\!\{ (s_i, \overset{\dashrightarrow}{s_is_j}, s_j) | s_i, s_j\!\in\!\bar{\mathcal{N}}\!\wedge\!(s_i, \cdot, s_j)\!\notin\!\mathcal{E} \\ 
\wedge [\![ s_i\!\rightarrow\!s_j ]\!]\!\wedge\!\omega(s_j)\!-\!\omega(s_i)\!\geq\!\epsilon \}
\end{aligned}
\end{equation*}
where $\bar{\mathcal{N}}$ represents the nodes on the shortest path between the source node $s_0$ (\ie empty solution) and the sink node $s_{|\mathcal{N}|}$ (\ie final solution) in $\mathcal{N}$, $[\![ s_i\!\rightarrow\!s_j ]\!]$ indicates that $s_j$ is reachable from $s_i$. 
Since the instruction of each shortcut edge does not inherently exist in the trajectory, for the two non-adjacent nodes connected by the shortcut, we create a pseudo instruction \(\overset{\dashrightarrow}{s_is_j}\) that effectively links the two non-adjacent nodes using a standard self-instruction mechanism~\cite{wang-etal-2023-self-instruct}. This involves creating an instruction by comparing two distinct codes.
This mechanism only extracts informative shortcuts on the shortest path and integrates compilation signals, naturally mitigating redundant solution backtracking and unexpected correct-to-failure degeneration.

\noindent\paragraph{Experience Gathering} To leverage the heuristically-discovered shortcuts identified from historical trajectories as past experiences, the agents accumulate their own experience pools with key-value pairs for the future reasoning:
\begin{equation*}
\begin{aligned}
\mathsf{S}_I\!= \bigcup_{\forall t \in \mathcal{T}} \{(s_i, \overset{\dashrightarrow}{s_is_j})  | (s_i, \overset{\dashrightarrow}{s_is_j}, s_j)\!\in\!\mathsf{S}_t \} \\
\mathsf{S}_A\!= \bigcup_{\forall t \in \mathcal{T}} \{(\overset{\dashrightarrow}{s_is_j}, s_j)  | (s_i, \overset{\dashrightarrow}{s_is_j}, s_j)\!\in\!\mathsf{S}_t \}
\end{aligned}
\end{equation*}
where $\mathsf{S}_t$ denotes the shortcut set extracted from a task $t$ and $\mathcal{T}$ is the whole training set.
This signifies that the instructor retains solution-to-instruction experiences to refine its instructional capabilities, while the assistant preserves instruction-to-solution experiences to enhance solution generation.

The design of shortcuts-as-experiences allows for an escape from the solution optimization capabilities that agents have already possessed in each step of their historical execution process, providing the possibility for more efficient shortcut-driven "accelerated" reasoning.

\subsection{Co-Reasoning}

The co-reasoning module is designed to combine the collective experience pools of agents, enabling communication through augmented instructions and solutions. By leveraging their respective experiential knowledge, these agents access and generate more refined answers, enhancing their collaborative task-solving abilities on unseen tasks.

The process begins with the instructor, armed with a solution-to-instruction memory \( \mathsf{S}_I \), encountering the current task solution \( s_j \). It starts by using a retrieval tool to access experiential instructions that closely match the latent meaning of the query~\cite{lewis2021retrievalaugmented}. 
These instructions serve as few-shot examples~\cite{zhao2023expel,rubin2021learning,min2022rethinking}, guiding the instructor's reasoning to produce \(i_{j+1}^*\), which is then shared with the assistant.
The assistant, equipped with an instruction-to-solution memory \( \mathsf{S}_A \), retrieves optimal solutions based on the received instruction. 
These solutions form the foundation for the assistant's few-shot examples, culminating in the formulation of a new solution \( s_{j+1}^* \).
This entire procedure can be represented as:
\begin{equation*}
\begin{aligned}
i_{j+1}^*\!=\!\mathbb{I}(s_j, \Bbbk(s_j, \mathsf{S}_I)) \ \  s_{j+1}^*\!=\!\mathbb{A}(s_j, \Bbbk(i_{j+1}^*, \mathsf{S}_A)) \\
\end{aligned}
\end{equation*}
where \( \Bbbk(q, s) \) denotes the retrieval of top-k matched results using \( q \) as a query in a key-value database \( s \), \( \mathbb{I} \) and \( \mathbb{A} \) are the in-context reasoning functions of the instructor and assistant, respectively, utilizing few-shot examples.

In each communication, the solution obtained is used as the next step in the agents' ongoing communication. This process for each task is represented as a sequence of pairs $\{(i_1^*, s_1^*), (i_2^*, s_2^*), \cdots\}$, where each pair includes an experience-enhanced instruction and the corresponding solution.

\section{Evaluation}

\noindent \paragraph{Baselines} We chose different types of LLM-driven software development paradigms as our baselines, which include both single-agent and multi-agent methodologies.
GPT-Engineer~\cite{GPTEngineer} is a foundational single-agent approach in the evolving domain of LLM-powered software agents; its standout feature is its exceptional proficiency in accurately comprehending input task requirements and employing one-step reasoning, which significantly enhances its efficiency in producing comprehensive software solutions at the repository level.
MetaGPT~\cite{hong2023metagpt} is an innovative framework that assigns diverse roles to various LLM-powered agents and incorporates standardized operating procedures to facilitate agent collaboration in software development; within each substep, solutions are generated through a single-step solution by agents with varying capabilities.
ChatDev~\cite{qian2023communicative} is an LLM-powered agent collaborative software development framework that organizes the entire software development process into waterfall-style phases (\eg code completion, code review, and system testing); within this framework, software agents engage in task-oriented and multi-turn communications that play a pivotal role in enhancing software development quality by iteratively providing instructions and solutions during their communications.

\noindent \paragraph{Datasets} We leveraged the SRDD dataset~\cite{qian2023communicative}, which contains various software requirement descriptions. This dataset, reflecting categories from major software store platforms, was crafted to minimize redundancy while enhancing originality and diversity. 
It consists of 1,200 software requirements, systematically arranged into 5 primary categories ( \texttt{Education}, \texttt{Work}, \texttt{Life}, \texttt{Game}, and \texttt{Creation}). These main categories are segmented into 40 distinct subcategories, with each subcategory containing 30 unique tasks.

\begin{table*}[t]
\centering
\resizebox{0.90\textwidth}{!}{
\begin{tabular}{lcccccc}
\toprule[1.5pt]
\multirow{1}{*}{\textbf{Method}} & \multirow{1}{*}{\textbf{Paradigm}}& \multicolumn{1}{c}{\multirow{1}{*}{\textbf{Completeness}}} & \multicolumn{1}{c}{\multirow{1}{*}{\textbf{Executability}}} & \multicolumn{1}{c}{\multirow{1}{*}{\textbf{Consistency} }} & \multicolumn{1}{c}{\multirow{1}{*}{\textbf{Quality}}} & \multicolumn{1}{c}{\multirow{1}{*}{\textbf{Duration} \begin{small}(s)\end{small}}}\\
\midrule[0.75pt]
GPT-Engineer & \includegraphics[height=10pt]{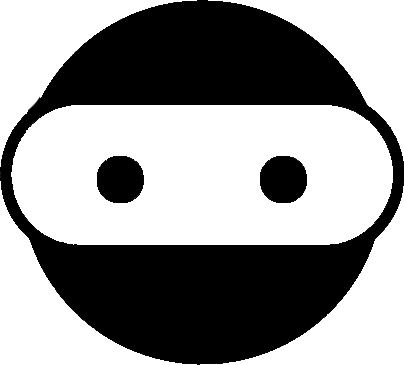}&0.4824 &0.3583 &0.7887&0.1363&\textbf{15.6000}\\
MetaGPT &\includegraphics[height=10pt]{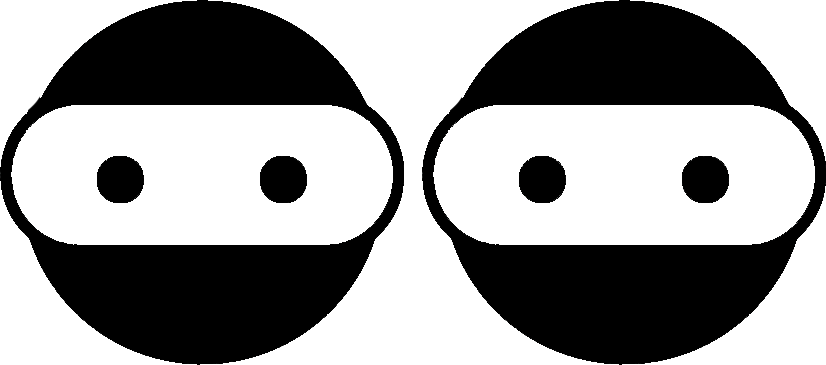}  &0.4472 & 0.4208&0.7649&0.1439& 154.0000\\
ChatDev &\includegraphics[height=10pt]{figs/p5.png}  &\underline{0.6131}&\underline{0.8800} &\underline{0.7909}&\underline{0.4267}&148.2150\\
\midrule[0.25pt]
Co-Learning & \includegraphics[height=10pt]{figs/p5.png} &\textbf{0.9497}&\textbf{0.9650} &\textbf{0.7970}&\textbf{0.7304}& \underline{122.7750}\\
\bottomrule[1.5pt]
\end{tabular}
}
\caption{Overall performance of the representative software development methods, encompassing both single-agent (\includegraphics[height=8pt]{figs/p4.png}) and multi-agent (\includegraphics[height=8pt]{figs/p5.png}) paradigms. Performance metrics are averaged across all tasks in the test set. The \textbf{highest} scores are formatted in bold, while the \underline{second-highest} scores are underlined.}
\label{tab:main-results}
\end{table*}

\noindent \paragraph{Metrics} Evaluating software is a challenging task, especially when trying to assess it on a holistic level. Traditional metrics like function-level code evaluation (e.g., \texttt{pass@k}) cannot seamlessly transfer to a comprehensive evaluation of entire software systems. This is primarily because it's often impractical to create manual or automated test cases for much of the software, particularly when dealing with complex interfaces, frequent communications, or non-deterministic feedback. As a solution, we use three quantifiable and objective dimensions to assess specific aspects of the software, and then combine these dimensions into a comprehensive metric for a more holistic evaluation:
\begin{enumerate}[$\bullet$]
\item \textit{Completeness} measures the software's ability to fulfill code completion in software development, quantified as the percentage of software without any "\texttt{TODO}" code snippets. A higher score indicates a higher probability of automated execution.  
\item \textit{Executability} assesses the software's ability to run correctly within a compilation environment, quantified as the percentage of software that compiles successfully and can run directly. A higher score indicates a higher probability of successful execution.
\item \textit{Consistency} evaluates the alignment between the generated software and the original natural language requirements. It is quantified as the cosine distance between the embeddings of the text requirements and the source code. A higher score indicates a greater degree of compliance with the requirements.
\item \textit{Quality} is a comprehensive metric that integrates the aspects of completeness, executability, and consistency to assess the overall quality of software.\footnote{To prevent over-complication, the quality metric is defined through the multiplication of completeness, executability, and consistency. Employing a simple average sum would produce similar results and conclusions.} A higher     quality score suggests a higher overall quality of the software generated, implying a lower need for further manual intervention.
\end{enumerate}

\noindent \paragraph{Implementation Details} Our system explicitly encompasses essential phases such as code complete, code review, and system testing.
In the co-tracking phase, we integrate \textit{GPT-3.5-Turbo} as the foundational model, limiting communication rounds between agents to a maximum of \textit{5} per phase.
For co-memorizing, we select \textit{text-embedding-ada-002} as the semantic embedder due to its exceptional performance in both text and code embeddings. We utilize \textit{MD5} as our hashing function, and \textit{Python-3.11.4} is employed to provide environment feedback. 
Agents have access to relevant tools like code checkers and compilers.
Additionally, it applies an information gain threshold of \textit{0.90} to retain informative shortcuts.
In the co-reasoning module, this approach selects the highest-ranked result from code and text experience pools as the in-context example for reasoning.
Besides, we divided the dataset into training, validation, and testing sets in a 4:1:1 ratio, using random hierarchical sampling to maintain a balanced category split.
The training set is utilized for co-tracking and co-memorizing to gather experiences, the validation set for choosing hyperparameters, and the test set for co-reasoning.
To maintain comparability in experimental results, all baseline models use identical parameters and environment settings.

\subsection{Quality Analysis}
\begin{figure}[t]
\centering
\centering
\includegraphics[width=\linewidth]{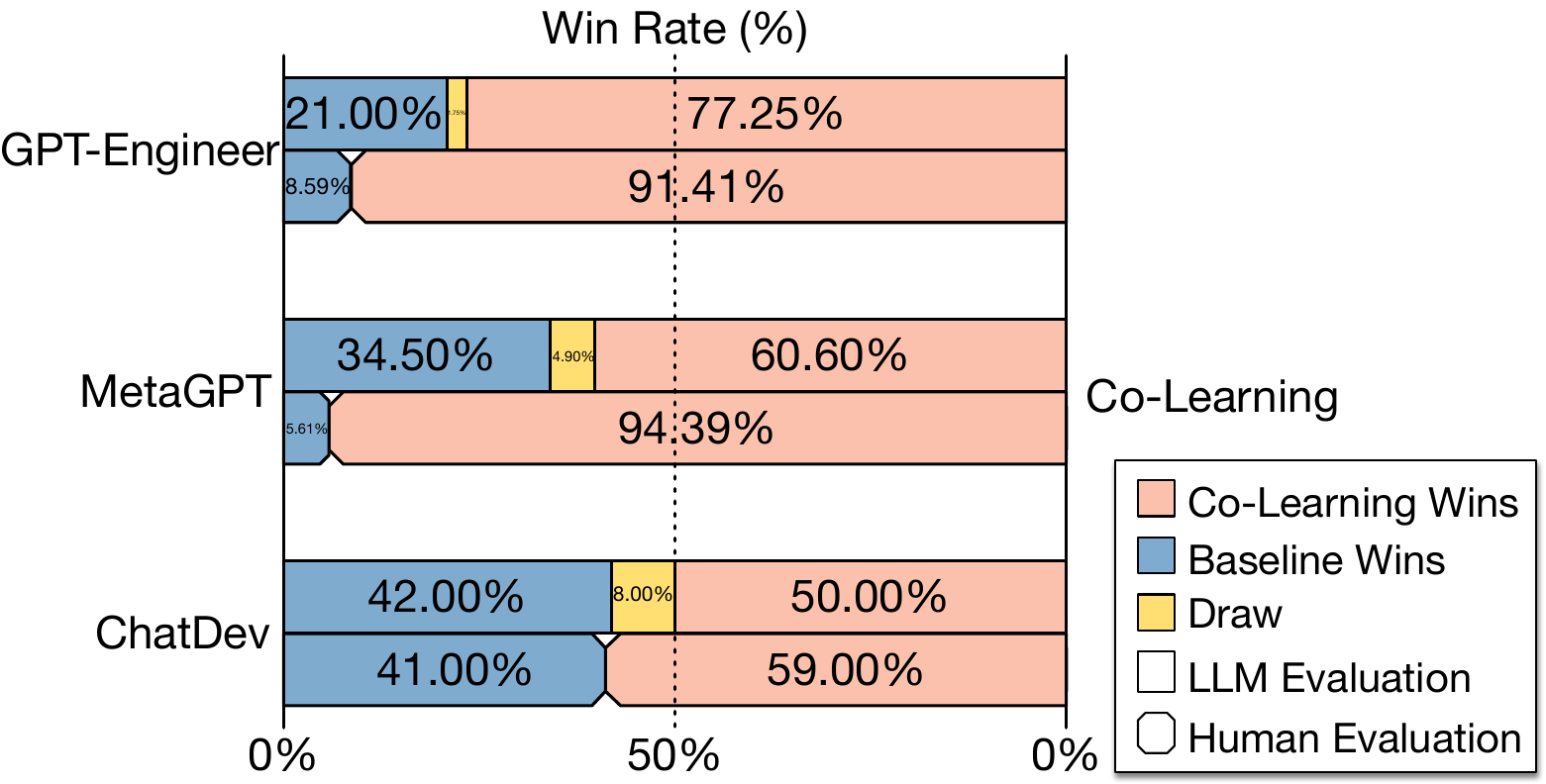}
\caption{GPT-4 and human as the evaluators for pairwise comparisons of the solutions generated.}
\label{fig:pairwise}
\end{figure}

\begin{figure*}[t]
    \centering
    \includegraphics[width=0.95\textwidth]{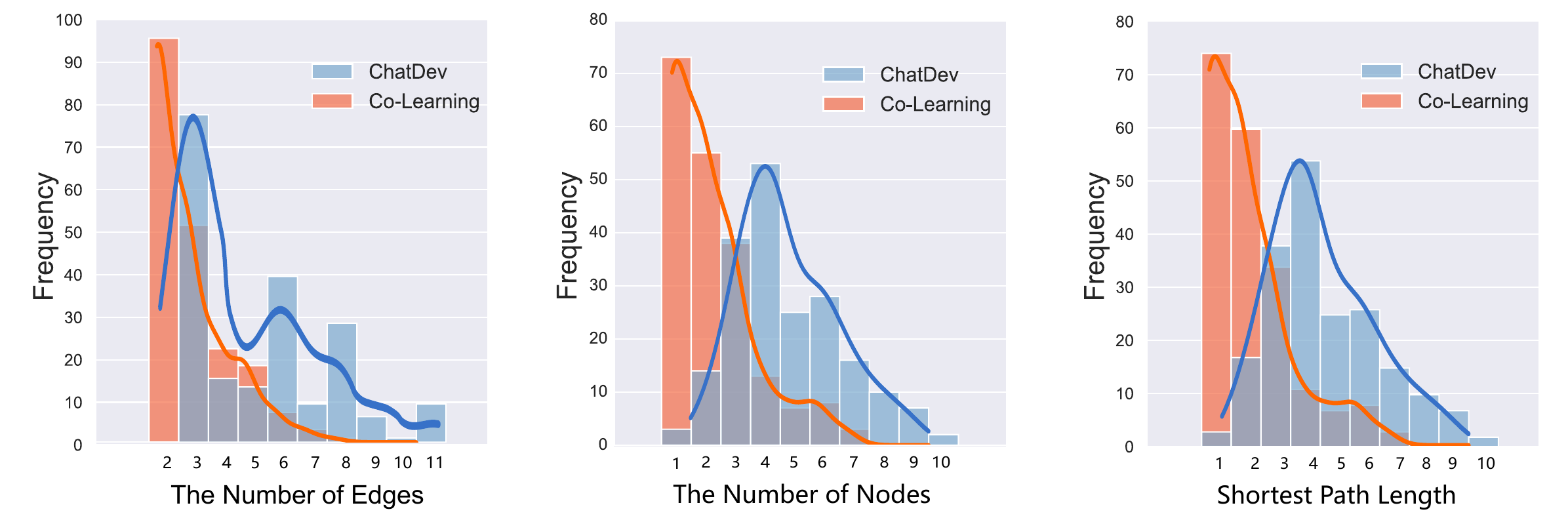}
    \caption{Distribution of key elements in task execution graphs. In a task execution graph, the number of edges represents the total communications between agents during code iterations in software development. The nodes count reflects the unique software source codes after hash deduplication, indicating the solution space during software optimization. The shortest path length shows the main path length in the task development process, excluding cycles and invalid attempts.}
    \label{fig:graph}
\end{figure*}

According to Table~\ref{tab:main-results}, our method, abbreviated as Co-Learning, significantly outperforms all three established baseline models in terms of quality.
The comparison with ChatDev, a powerful multi-agent framework, is particularly striking, as Co-Learning significantly boosts the comprehensive metric from 0.4267 to 0.7304, highlighting the effectiveness of utilizing agents' past experiences in addressing unseen tasks.
The efficacy of our method primarily stems from the agents' proficiency in recalling and applying high-quality shortcuts from past trajectories, leading to notable enhancements in key performance metrics like completeness and executability for unseen tasks.

Moreover, Co-Learning outperforms GPT-Engineer, illustrating the superiority of the multi-agent approach in segmenting task-solving into discrete subtasks, in contrast to a single-step strategy. Through active communication, each agent contributes to a dynamic collaboration, steering towards cohesive and automated solutions for task execution. The effectiveness of this approach in task decomposition is further validated by contemporary studies \cite{wei2022chain}. Additionally, although Co-Learning requires more time compared to single-agent methods, it proves to be more time-efficient than other multi-agent approaches, suggesting that while multi-agent communications inherently take longer, the strategic use of "shortcut" patterns from past experiences effectively reduces reasoning time, striking a balance between performance and duration.

Upon closer examination, while Co-Learning exhibits notable strengths in the areas of completeness and executability, it demonstrates only a slight enhancement in the aspect of consistency.
This result could likely stem from the embedding models' broad-grained semantic representation capabilities for text and code, which might not be adequately sensitive to discern extremely nuanced inconsistencies.
This discovery presents an exciting research opportunity to develop advanced criteria and metrics for assessing software's consistency with its text requirements, emphasizing the need for more refined evaluation methodologies.

To further validate the efficacy of our method from an alternative perspective, we drew inspiration from previous work~\cite{li2023camel}, which adopts both GPT-4 and human participants for pairwise comparisons of the solutions producted by agents.\footnote{For each task, GPT-4 evaluation compares solutions from both methods, avoiding positional bias~\cite{wang2023largelanguage}. In the human evaluation, thirty computer science researchers assessed the software, with solutions from both methods randomly presented for their preference-based selection.}
Figure \ref{fig:pairwise} illustrates that our Co-Learning method consistently surpasses other baseline methods, as evidenced by its higher average win rates in both evaluations involving GPT-4 and human participants.
The results also reveal that ChatDev consistently demonstrates a notably high win rate compared to other baseline methods, indicating its strength as a robust baseline. This success is largely due to ChatDev's approach of performing detailed task decomposition and software optimization via multi-round agent communications, effectively addressing potential shortcomings. 
The Co-Learning approach, akin to a ChatDev variant emphasizing experiential agents, highlights the importance of agent experience accumulation.
\subsection{Efficiency Analysis}
Recall that we explicitly construct a task execution graph in which shortcuts are heuristically extracted as agents' experiences.
In this section, we delve deeper into the graph analysis to uncover fundamental patterns in the task execution process by agents. For comparison purposes, we selected ChatDev as it represents the strongest baseline currently available.
Figure~\ref{fig:graph} shows that, compared to ChatDev, Co-Learning trends towards fewer numbers in terms of the number of edges, nodes, and the length of the shortest path.
This suggests a decrease in the number of iterations required for software development and a simplification of the solution space during software optimization.
The enhancement in efficiency can be largely credited to the strategic utilization of shortcuts linking non-adjacent nodes in the graph, which enables agents to leverage previous task execution experiences while simultaneously boosting their future task-solving skills more effectively through the adoption of "shortcut thinking".
In conjunction with Table~\ref{tab:main-results} and Figure~\ref{fig:pairwise}, this evidence demonstrates that the Co-Learning method streamlines the software development process by decreasing unnecessary iterations, thereby not only boosting overall efficiency but also delivering higher quality solutions with fewer agent communication.

\subsection{Effectiveness Analysis}
In this part, we delve into the distinct roles of instructors and assistants within the agent-collaboration framework, focusing on scenarios where either one agent alone has past experiences or both agents are inexperienced.
As illustrated in Table~\ref{tab:abla1}, relying solely on a single agent leads to a marked decline in overall performance, manifesting in an increase in execution communications and a decrease in the quality.
Furthermore, it is notable that systems with an experienced instructor perform worse than those with an experienced assistant (0.5305 vs. 0.6840), indicating the greater significance of the assistant's role in task execution, despite both roles being essential.
In situations where neither agent type has experience, the system reverts to its traditional technological capabilities, resulting in the least effective performance solutions.
These findings highlight the necessity for both instructors and assistants to be experienced; systems lacking this or relying on only one experienced agent demonstrate diminished execution efficiency and quality. 

To further confirm the effectiveness of the key mechanisms employed in our framework, we conduct experiments using three different configurations and the results are presented in Table~\ref{tab:abla1}, consistently exhibiting a decline in performance.
The \textit{adjacent-execution} variant adopts the experiences of adjacent nodes from the original trajectories (after successfully compilation) rather than utilizing shortcuts, equivalent to ExpeL~\cite{zhao2023expel}. 
This approach leads to a significant increase in the number of experiences, ultimately resulting in a decrease in the quality of experiences. This, in turn, validates the effectiveness of the proposed "shortcuts-as-experiences" scheme.
The \textit{longest-shortcut-only} variant discards all intermediate shortcuts and exclusively relying on the "longest" shortcut that directly connects the start and end nodes. It reveals the underlying models' limitations and its inability to handle new tasks using only single-step historical experiences, indicating an overstretch of the agent's contextual reasoning ability.
The \textit{graph-unconstructed} variant extracts shortcuts directly from the original trajectory without constructing graphs.
The obtained solution also reveals that over-dependence on past experience fails to achieve optimal performance, as it creates many shortcuts for possible graph nodes, leading to unnecessary repetition in the test set and decreasing task execution efficiency.
This result confirms our observation that the progression of arbitrary adjacent nodes does not necessarily lead to continuously improved solutions.
This emphasizes the significance of using heuristic shortcuts in deduplicated graphs to strike a balance between quantity and quality.

\begin{table}[t]
\centering
\resizebox{\linewidth}{!}{
\begin{tabular}{lcccc}
\toprule[1.5pt]
\multicolumn{1}{c}{\textbf{Method}} & \multicolumn{1}{c}{\textbf{\#Experiences}} & \multicolumn{1}{c}{\textbf{\#Nodes}}  & \multicolumn{1}{c}{\textbf{\#Edges}} & \multicolumn{1}{c}{\textbf{Quality}} \\
\midrule[0.75pt]
Co-Learning & (537, 537) & \textbf{2.3100} & \textbf{3.0100} & \textbf{0.7304} \\
\midrule[0.25pt]
\quad $\diagdown$Instructor's Experiences & (0, 537) & 3.3500 & 3.8850 & 0.6840 \\
\quad $\diagdown$Assistant's Experiences & (537, 0) & 4.4422 & 5.0352 & 0.5305\\
\quad $\diagdown$Both Experiences & (0, 0) & 3.9450 & 4.7950 & 0.4267 \\
\midrule[0.25pt]
\quad $\circlearrowleft$Adjacent-Execution & (1604, 1604) & 3.7000 & 4.5000 & 0.6398 \\
\quad $\circlearrowleft$Longest-Shortcut-Only & (332, 332) & 2.8700 & 3.5200 & 0.6752 \\
\quad $\circlearrowleft$Graph-Unconstructed & (605, 605) & 2.7000 & 3.4350 & 0.6821 \\
\bottomrule[1.5pt]
\end{tabular}
}
\caption{Ablation study on main roles or mechanisms of our framework. $\circlearrowleft$ and $\diagdown$ denote the replacement operation and the removing operation respectively. ($a$, $b$) indicates that the instructor and assistant are equipped with $a$ and $b$ heuristic shortcuts respectively.}
\label{tab:abla1}
\end{table}

\subsection{Sensitivity Analysis}
In this section, we explores the effects of two key parameters in the co-reasoning process: $\mathsf{k}$ (the number of matched results in retrieval) and $\theta$ (the semantic similarity threshold utilized in retrieval). The results obtained from retrieval come in two forms: text and code, so the number of matched results are denoted as \(\mathsf{k}_\text{text}\) and \(\mathsf{k}_\text{code}\), ranging from 1 to 5, and the semantic similarity thresholds are denoted as \(\theta_\text{text}\) and \(\theta_\text{code}\), ranging from 0.0 to 1.0 in increments of 0.20.

\begin{figure}[t]
    \centering
    \includegraphics[width=0.9\linewidth]{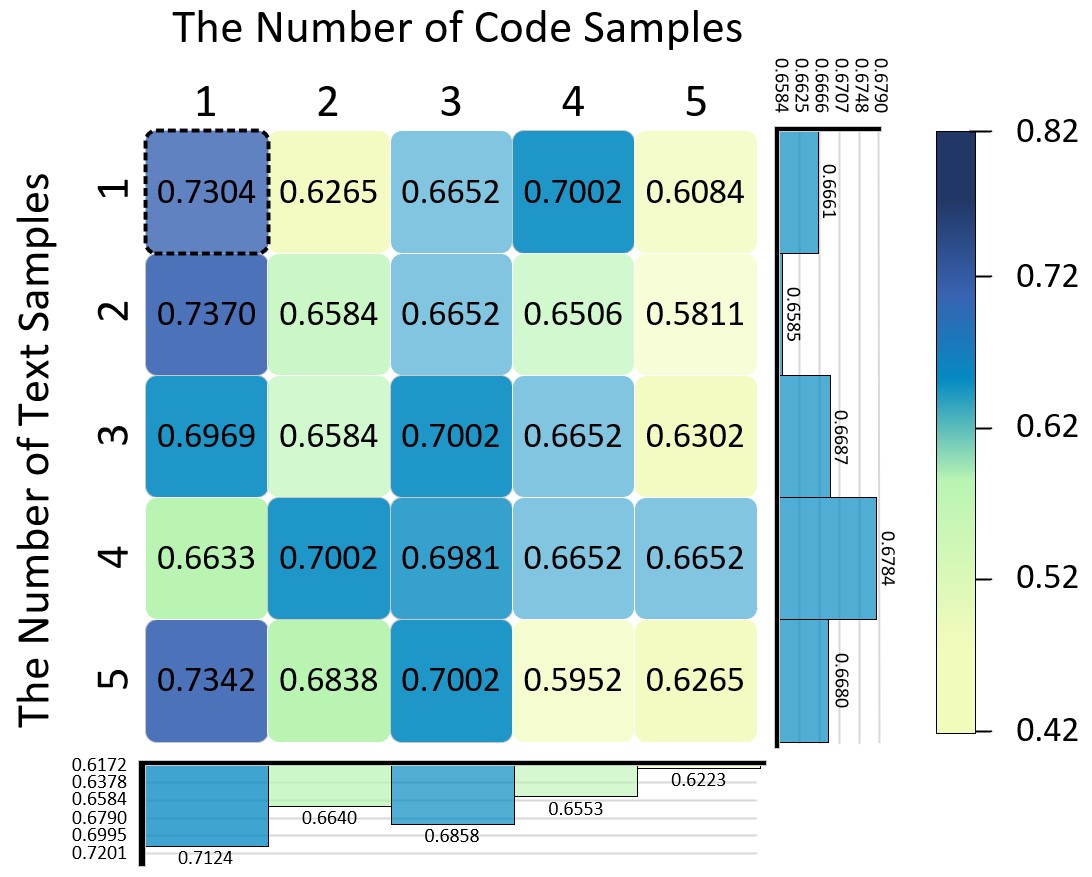}
    \caption{The effect of the top-k, with the dashed line indicating the default configuration utilized in the above experiments.}
    \label{fig:topK}
\end{figure}

\begin{figure}[t]
    \centering
    \includegraphics[width=0.84\linewidth]{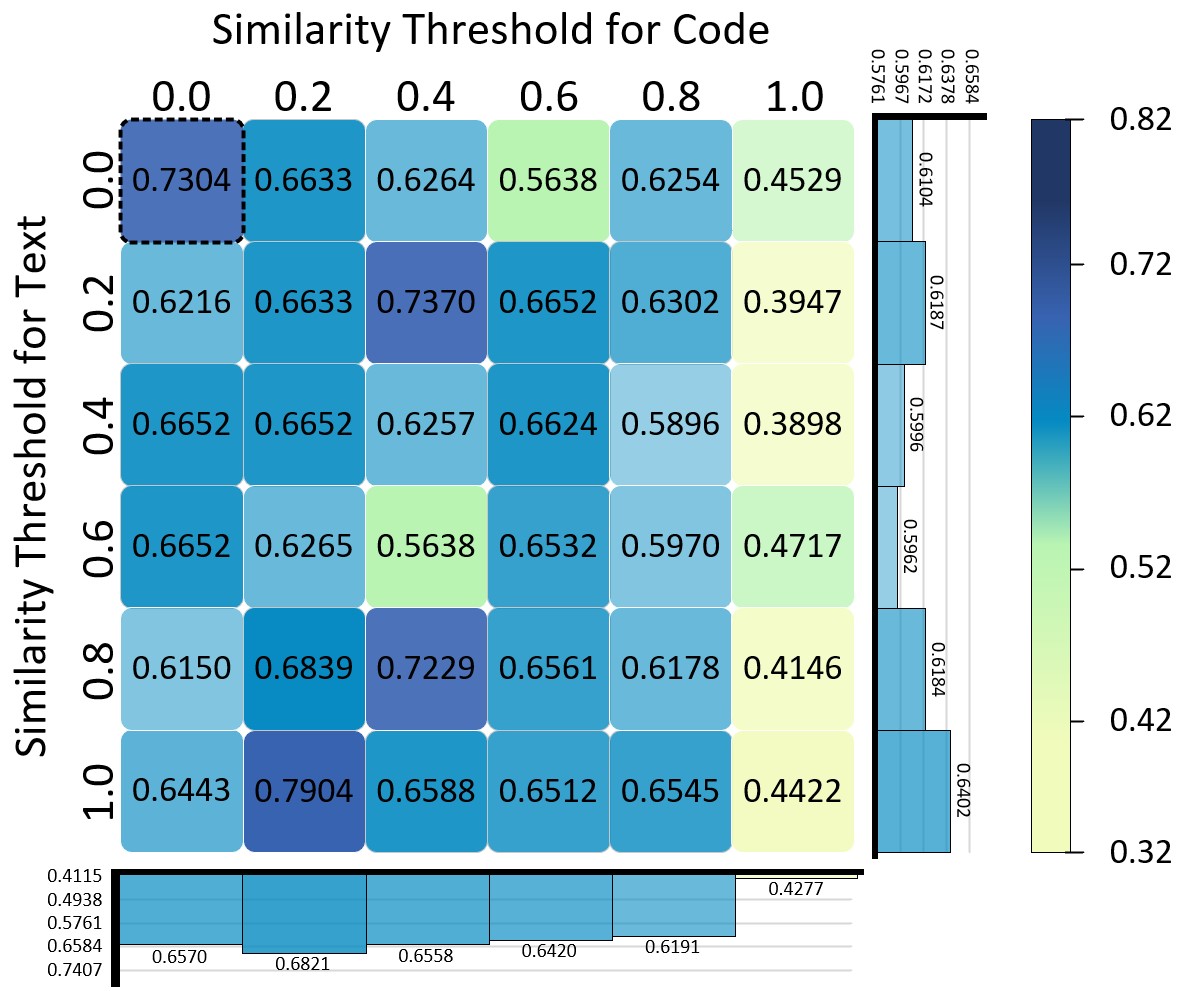}
    \caption{The effect of the semantic similarity threshold, with the dashed line indicating the default configuration utilized in the above experiments.}
    \label{fig:theta}
\end{figure}

Figure \ref{fig:topK} shows that the best retrieval performance is achieved under the configuration \((\mathsf{k}_\text{code} = 1, \mathsf{k}_\text{text} = 2)\), surpassing the default setting \((\mathsf{k}_\text{code} = 1, \mathsf{k}_\text{text} = 1)\).
This indicates that although the proposed framework has already achieved success under its default setting, there is still potential for further improvement through hyperparameter optimization.
Additionally, Figure \ref{fig:topK} suggest that optimizing hyperparameters related with code yields superior outcomes than optimizing hyperparameters related with text, aligning well with the conclusions from our previous experiments.
Figure \ref{fig:theta} illustrates that the $\theta$ parameter has a limited impact on the results, which is attributed to the consistently high semantic similarity between the current query and the retrieval result, often surpassing 0.85.

\section{Conclusion}
Recognizing the absence of a mechanism for integrating cumulative experiences from past tasks in agent collaboration, we have proposed Experiential Co-Learning, a framework that encourages collaboration between autonomous agents. Through co-tracking, co-memorizing, and co-reasoning, this approach enables agents to efficiently handle unseen software development tasks by drawing on past experiences and providing mutual support.
The quantitative analysis effectively showcased significant improvements in quality, leading to reduced execution times, decreased repetitive errors, and a decreased reliance on additional human intervention.
We anticipate that our insights will initiate a paradigm shift in shaping the design of multi-agent collaboration, propelling agents towards achieving greater autonomy and contributing to their evolutionary growth in cooperative learning.

\section{Limitations}
Our study has explored the capabilities of cooperative autonomous agents, yet both researchers and practitioners should be mindful of certain limitations and risks.
Firstly, the ability of autonomous agents to produce software might be overestimated. In co-learning, autonomous agents still tend to implement the simplest logic in absence of necessary and clear requirements, indicating that these technologies are more suitable for prototype systems rather than real-world applications.
Secondly, unlike traditional function-level code generation approach, automating the evaluation of general-purpose software is exceptionally challenging. Though recent efforts aimed to employ \textit{Human Revision Cost}~\cite{hong2023metagpt}, manual verification is still impractical. This paper instead focuses on three objective and crucial dimensions and a comprehensive quality. Future research need take additional dimensions like robustness into consideration.

\section*{Acknowledgments}
The work was supported by the National Key R\&D Program of China (No.2022ZD0116312), the Postdoctoral Fellowship Program of CPSF under Grant Number GZB20230348, and Tencent Rhino-Bird Focused Research Program.

\bibliography{references}
\end{document}